\documentclass[10pt,twocolumn,letterpaper]{article}

\usepackage{iccv}
\usepackage{times}
\usepackage{epsfig}
\usepackage{graphicx}
\usepackage{amsmath}
\usepackage{amssymb}

\usepackage{bbm}
\usepackage{calc}
\usepackage[table,xcdraw]{xcolor}
\usepackage{xcolor}
\usepackage{subfigure}
\usepackage{array}
\usepackage{cases}
\usepackage{booktabs}
\usepackage{multirow}
\usepackage{colortbl}
\usepackage[ruled]{algorithm}
\usepackage{algpseudocode}
\usepackage{lipsum}

\newcommand{\eqnref}[1]{Eq.~\eqref{#1}}

\newcommand{\proposed}{PnP-GA}
\definecolor{mygray}{gray}{.9}

\newcommand{\rev}[1]{\textcolor[rgb]{0.0,0.0,0.0}{#1}}  	  
\newcommand{\mydarkred}[1]{\textcolor[rgb]{0.8,0.0,0.0}{ #1}}
\newcommand{\mydarkblue}[1]{\textcolor[rgb]{0.2,0.5,0.8}{ #1}}
\newcommand{\mydarkgreen}[1]{\textcolor[rgb]{0.7,0.7,0.25}{ #1}}

\newcommand{\upscore}[1]{\footnotesize{\mydarkred{$\blacktriangledown$ #1\%}}}

\definecolor{color1}{rgb}{0.22,0.45,0.70}  
\definecolor{color2}{rgb}{0.45,0.45,0.45}  

\usepackage[pagebackref=true,breaklinks=true,letterpaper=true,colorlinks,bookmarks=false]{hyperref}

\iccvfinalcopy 

\ificcvfinal\fi

\begin{document}

\title{Generalizing Gaze Estimation with Outlier-guided Collaborative Adaptation}

\author{Yunfei~Liu\textsuperscript{\rm 1, $\dag$}\qquad~Ruicong~Liu \textsuperscript{\rm 1, $\dag$}\qquad~Haofei Wang\textsuperscript{\rm 2}\qquad~Feng~Lu\textsuperscript{\rm 1, 2, }\thanks{ Corresponding Author. $\dag$ denotes equal contribution.
		    }\\
		{\textsuperscript{\rm 1} State Key Laboratory of VR Technology and Systems, 
		School of CSE, Beihang University}  \\
		{\textsuperscript{\rm 2} Peng Cheng Laboratory, Shenzhen, China} \\
		\small{\texttt{\{lyunfei, liuruicong, lufeng\}@buaa.edu.cn}} \qquad
		\small{\texttt{ wanghf@pcl.ac.cn}}
	}

\maketitle
\ificcvfinal\thispagestyle{empty}\fi

\begin{abstract}   
	Deep neural networks have significantly improved appearance-based gaze estimation accuracy. However, it still suffers from unsatisfactory performance when generalizing the trained model to new domains, e.g., unseen environments or persons.  In this paper, we propose a plug-and-play gaze adaptation framework (\proposed), which is an ensemble of networks that learn collaboratively with the guidance of outliers. Since our proposed framework does not require ground-truth labels in the target domain, the existing gaze estimation networks can be directly plugged into \proposed~ and generalize the algorithms to new domains. We test \proposed~ on four gaze domain adaptation tasks, ETH-to-MPII, ETH-to-EyeDiap, Gaze360-to-MPII, and Gaze360-to-EyeDiap. The experimental results demonstrate that the \proposed~ framework achieves considerable performance improvements of 36.9\%, 31.6\%, 19.4\%, and 11.8\% over the baseline system. The proposed framework also outperforms the state-of-the-art domain adaptation approaches on gaze domain adaptation tasks. Code has been released at \url{https://github.com/DreamtaleCore/PnP-GA}.
\end{abstract}

\section{Introduction}
	
	Eye gaze is an important indicator of human attention. It has been exploited in various applications, such as human-robot interaction~\cite{A:admoni2017social,A:terziouglu2020designing,wang2015hybrid}, virtual/augmented reality games~\cite{A:burova2020utilizing,A:konrad2020gaze,A:wang2020comparing}, intelligent cockpit systems~\cite{A:gerber2020self}, medical analysis~\cite{A:castner2020deep}, etc. With the development of deep learning techniques, appearance-based gaze estimation attracts much attention recently. To improve gaze estimation accuracy, many large-scale gaze estimation datasets have been proposed~\cite{G:funes2014eyediap,G:kellnhofer2019gaze360,G:zhang2020eth,G:zhang2017mpiigaze}. Due to the differences of subjects, background environments, and illuminations between these datasets, the performance of gaze estimation models that are trained on a single dataset usually dramatically degrades when testing on a new dataset.

	\begin{figure}[t]
		\begin{center}
			\includegraphics[width=\linewidth]{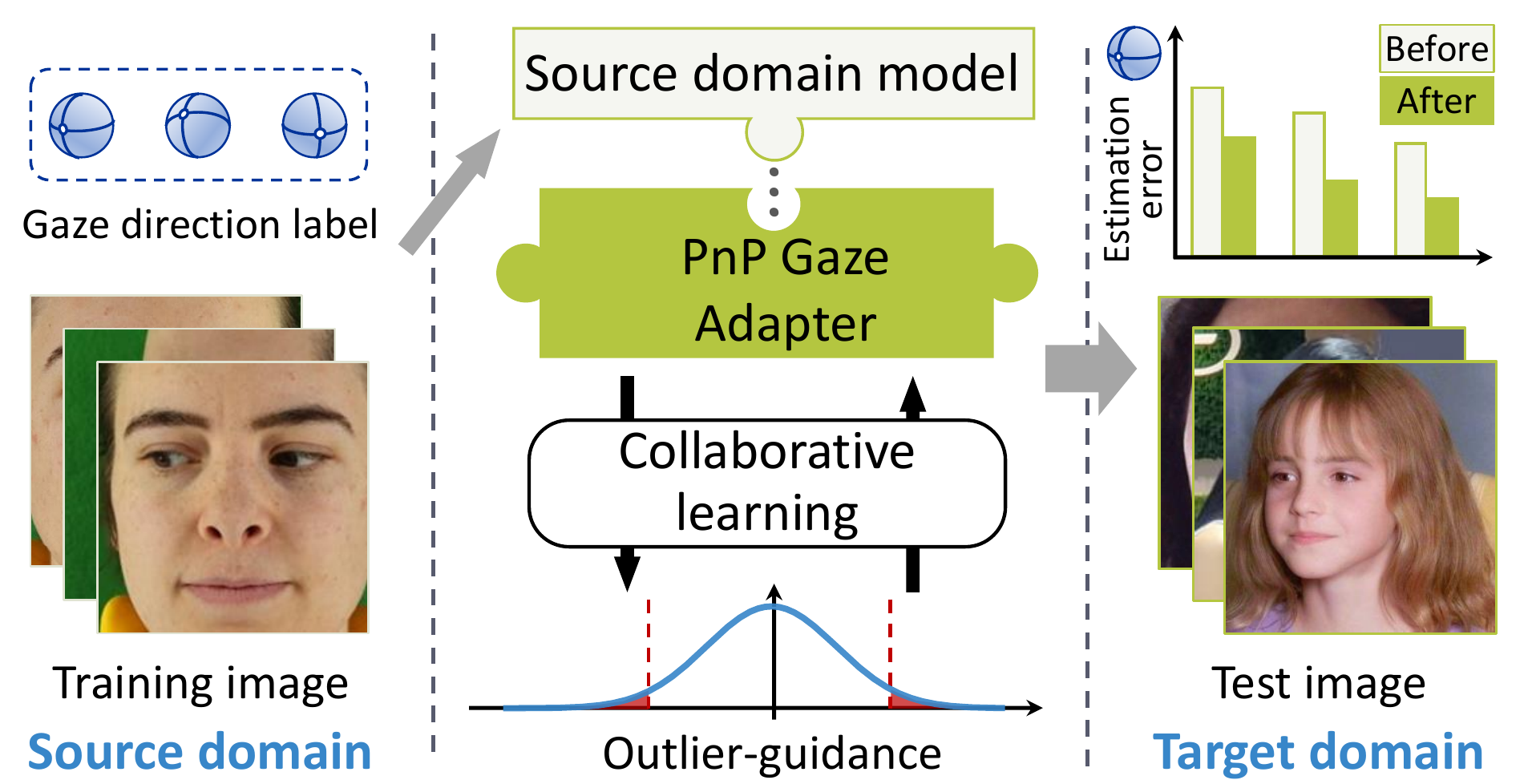}
		\end{center}
		\caption{Overview of the proposed Plug-and-Play (PnP) adaption framework for generalizing gaze estimation to a new domain.}
		\label{fig:teaser}
	\end{figure}
	
	To tackle such a gap in gaze estimation, several works have been proposed by utilizing adversarial training (ADV)~\cite{G:ashish2017adgaze,G:wang2019adgaze}, few-shot learning (FSL)~\cite{G:park2019fewshot,G:yu2019improvefewshot}, or embedding with prediction consistency~\cite{G:guo2020dagaze}. However, these works mainly focus on the inter-person gap (\ie, personal calibration) from the same dataset, where the environmental conditions are often very similar. When generalizing the gaze estimation models to new datasets, the differences between the image capture devices and the environments cannot be neglected, as illustrated in Fig.~\ref{fig:teaser}. Consequently, a gaze estimation model trained in the source	domain cannot achieve satisfactory performance by directly applying to the target domain. In this paper, we focus on domain adaptation between different datasets, which is more common in practical applications.
	
	Unlike most of the existing works in gaze domain adaptation following supervised learning paradigm, this work aims to generalize the gaze estimation in an unsupervised manner. This is because we usually do not have access to the ground-truth labels in a new domain. The task can be viewed as an unsupervised domain adaptation (UDA) problem. Table~\ref{tab:diff} shows the differences between the existing gaze domain adaptation approaches and UDA. While the ADV and FSL approaches are model-specific, the UDA and fine-tuning approaches are model-agnostic. It means that these methods are capable of adapting arbitrary network architectures. However, the fine-tuning approach requires a number of target labels. The UDA approach is more challenging than conventional approaches since the lack of annotated data in a new domain prohibits fine-tuning of the pre-trained models. Many UDA methods for other computer vision tasks have explored to take additional priors, \eg, shape priors for hand segmentation~\cite{D:cai2020uma}, pseudo labels for person re-identification~\cite{D:ge2020mutual}, 
	\rev{random sampler~\cite{D:kuhnke2019deep} for head pose estimation,} 
	\etc. However, the aforementioned priors are not available on gaze estimation tasks, since gaze directions have no specific shapes or labels. 
    
	\begin{table}[htbp]
    	\begin{center}
    		\caption{Differences among the domain adaptation approaches.}
    		\label{tab:diff}
    		\setlength{\tabcolsep}{2mm}
    		{   \small
    			\begin{tabular}{lllc}
    				\bottomrule[1.2pt]
    				\specialrule{0em}{1pt}{1pt}
    				Methods & \# Target images & \# Target labels  & Model\\
    				\hline
    				\specialrule{0em}{1pt}{1pt}
    				Fine-tuning         		                        & \mydarkblue{$\blacksquare\blacksquare\blacksquare\blacksquare\blacksquare$}  & \mydarkblue{$\blacksquare\blacksquare\blacksquare\blacksquare\blacksquare$}  & \mydarkgreen{Agnostic} \\
    			    FSL~\cite{G:park2019fewshot,G:yu2019improvefewshot} & \mydarkblue{$\blacksquare$}         & \mydarkblue{$\blacksquare$} & \mydarkblue{Specific} \\
    			    ADV~\cite{G:ashish2017adgaze,G:wang2019adgaze}      & \mydarkblue{$\blacksquare\blacksquare$}  & \mydarkgreen{Not require} & \mydarkblue{Specific} \\
    			    UDA (Ours)                                          & \mydarkblue{$\blacksquare$}    & \mydarkgreen{Not require}  & \mydarkgreen{Agnostic} \\
    				\bottomrule[1.2pt]
    		\end{tabular}  }
    	\end{center}
    \end{table}

    In this paper, we propose an outlier-guided plug-and-play collaborative learning framework for generalizing gaze estimation with unsupervised domain adaptation. Optimizing the error-prone outliers has been proved to be helpful for model generalization~\cite{C:jiang2018mentornet,C:zhang2017understanding}. We follow the similar idea and take the outliers of pre-trained models' outputs in the target domain as noisy labels. We employ two groups of networks that collaboratively teach each other. While these two groups share the same architecture, one group is the momentum version of another group. The primary contributions of this paper are summarized as follow:
    
	\begin{itemize}
		\item We propose a plug-and-play gaze adaptation framework (\proposed) for generalizing gaze estimation in new domains. 
		Existing gaze estimation networks can be easily plugged into our framework without modification of their architectures.
		\item We propose an outlier-guided collaborative learning strategy for the unsupervised domain adaptation. It only requires a few samples from the target domain without any labels. We specifically design an outlier-guided loss to better characterize the outliers and guide the learning.
		\item The \proposed~framework shows exceptional performances with plugging of the existing gaze estimation networks. Our system achieves performance improvements over the baseline system of 36.9\%, 31.6\%, 19.4\% and 11.8\% on four gaze adaptation tasks: ETH-to-MPII, ETH-to-EyeDiap, Gaze360-to-MPII and Gaze360-to-EyeDiap.
	\end{itemize}

\section{Related Works}

\subsection{Gaze estimation}
Appearance-based gaze estimation methods have gradually become a research hotspot. Recent studies have significantly improved the gaze estimation accuracy on public datasets by using coarse-to-fine strategy\cite{G:cheng2020canet}, domain adaptation\cite{G:guo2020dagaze,G:ashish2017adgaze}, and adversarial learning approach\cite{G:wang2019adgaze}. 

However, Most studies investigate the person-specific adaptation of gaze estimators within the same dataset\cite{G:lu2014alr, G:park2019fewshot, G:yu2019improvefewshot}, the model generalization across the dataset is often ignored in previous studies. Such task is rather challenging due to the huge differences of the distribution between different datasets. Considering the diversity of the real-world environment, even if the model performs well within the dataset, it is difficult to apply to practical applications. More recently, ETH-XGaze\cite{G:zhang2020eth}, Gaze360\cite{G:kellnhofer2019gaze360}, \etc. large-scale gaze datasets with a wider distribution have been proposed, which have greatly promoted the development of gaze estimation to practical applications. Recent work \cite{G:zhang2020eth} has begun to consider the cross dataset generalization capabilities of gaze estimation methods. Here, we propose an unsupervised approach to improve the cross-dataset and cross-domain performance of gaze estimation methods.

\subsection{Unsupervised domain adaptation}
Unsupervised domain adaptation has been extensively explored in deep learning field. Previous unsupervised domain adaptation approaches can be categorized into two techniques: adversarial learning and self-training. Adversarial learning based approaches are motivated by generative adversarial networks(GAN)\cite{D:good2014gan}. A large number of such methods have been developed, such as ADDA\cite{D:tzeng2017adversarial}, CDAN\cite{D:long2018cdan}, CyCADA\cite{D:hoffman2017cycada}, Symnets\cite{D:zhang2019symnets}, GVBGD\cite{D:cui2020gradually}, \etc. The key idea is to make the domain discriminator and the generator play a min-max game, thereby it explicitly reduces the distance between the source domain and the target domain. Self-training based approaches are motivated by semi-supervised learning\cite{D:lee2013pseudo, D:antti2017meanteacher}, which usually trains the model using the reliable predictions of itself, such as Pseudo-label method\cite{D:lee2013pseudo}, Mean Teacher method\cite{D:antti2017meanteacher}, \etc. Recent works \cite{D:cai2020uma, D:chen2019maxiumsquare, D:french2018ensembling, D:zou2018classbalance} apply self-training to unsupervised domain adaptation, and achieve significant results. Compared with adversarial learning, although the self-training methods can only perform feature alignment implicitly, the adaptation performance is usually better.

\begin{figure}[t]
	\begin{center}
		\includegraphics[width=0.9\linewidth]{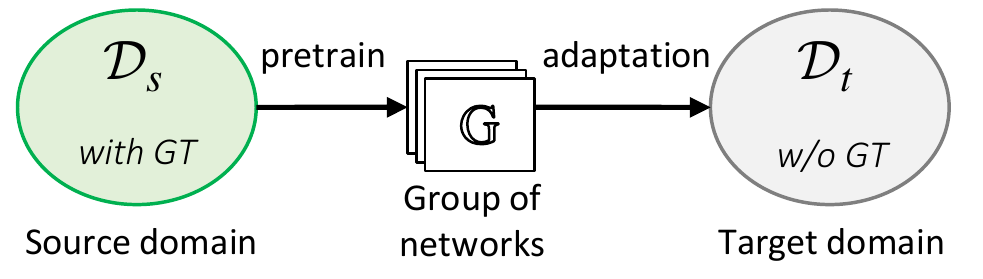}
	\end{center}
	\caption{Illustration of unsupervised domain adaptation for gaze estimation.}
	\label{fig:task-def}
\end{figure}

However, the majority of the previous domain adaptation works focus on classification tasks rather than regression tasks. For gaze domain adaptation, unsupervised domain adaptation is even difficult due to the continuity nature of gaze directions. Thus, there still remains much space to be explored.

\section{Task Definition} \label{sec:taskddef}

Fig.\ref{fig:task-def} illustrates the unsupervised domain adaptation for gaze estimation. We denote the source domain data as $\mathcal{D}_s = \{(\mathbf{x}^s_i, \mathbf{y}^s_i) |_{i=1}^{N_s}\}$, where $\mathbf{x}^s_i$ and $\mathbf{y}^s_i$ denote the \textit{i}-th training image and the corresponding gaze direction, $N_s$ is the number of images, $\mathbf{y}^s_i$ contains the pitch and yaw angle of the gaze direction. The target domain data is denoted as $\mathcal{D}_t = \{(\mathbf{x}^t_i) |_{i=1}^{N_t}\}$, where $N_t$ is the number of images, $\mathbf{x}^t_i$ is the \textit{i}-th image. It contains no ground-truth gaze direction.

Suppose we have a group of baseline gaze estimation models $\mathbb{G} = \{G(\centerdot|\theta_k)|_{k=1}^{H}\}$ with parameters $\theta_k$ learned using training data from the source domain $\mathcal{D}_s$, where $H$ is the number of models.
The pre-trained baseline models may not generalize to a target domain since the test samples from $\mathcal{D}_t$ and $\mathcal{D}_s$ usually have different distributions. Our task is to adapt the pre-trained models to the new domain $\mathcal{D}_t$ without requiring any ground-truth labels.

\noindent\textbf{Datasets.}
Four different datasets are employed as four different domains: ETH-XGaze ($\mathcal{D}_E$), Gaze360 ($\mathcal{D}_G$), MPIIGaze ($\mathcal{D}_M$) and EyeDiap ($\mathcal{D}_D$). The details of the datasets are as follows:

\textit{ETH-XGaze dataset} ($\mathcal{D}_E$)~\cite{G:zhang2020eth} is collected under adjustable illumination conditions and a custom hardware setup. 

\textit{Gaze360 dataset} ($\mathcal{D}_G$)~\cite{G:kellnhofer2019gaze360} 
consists of 238 subjects in indoor and outdoor environments. 

\textit{MPIIGaze dataset} ($\mathcal{D}_M$)~\cite{G:zhang2017mpiigaze} is collected from 15 subjects in real-world environments. Therefore it has various illumination conditions. 

\textit{EyeDiap dataset} ($\mathcal{D}_D$)~\cite{G:funes2014eyediap} consists of a number of video clips from 16 subjects, which are collected under screen targets or 3D floating balls. 

\begin{figure}[t]
	\begin{center}
		\includegraphics[width=0.95\linewidth]{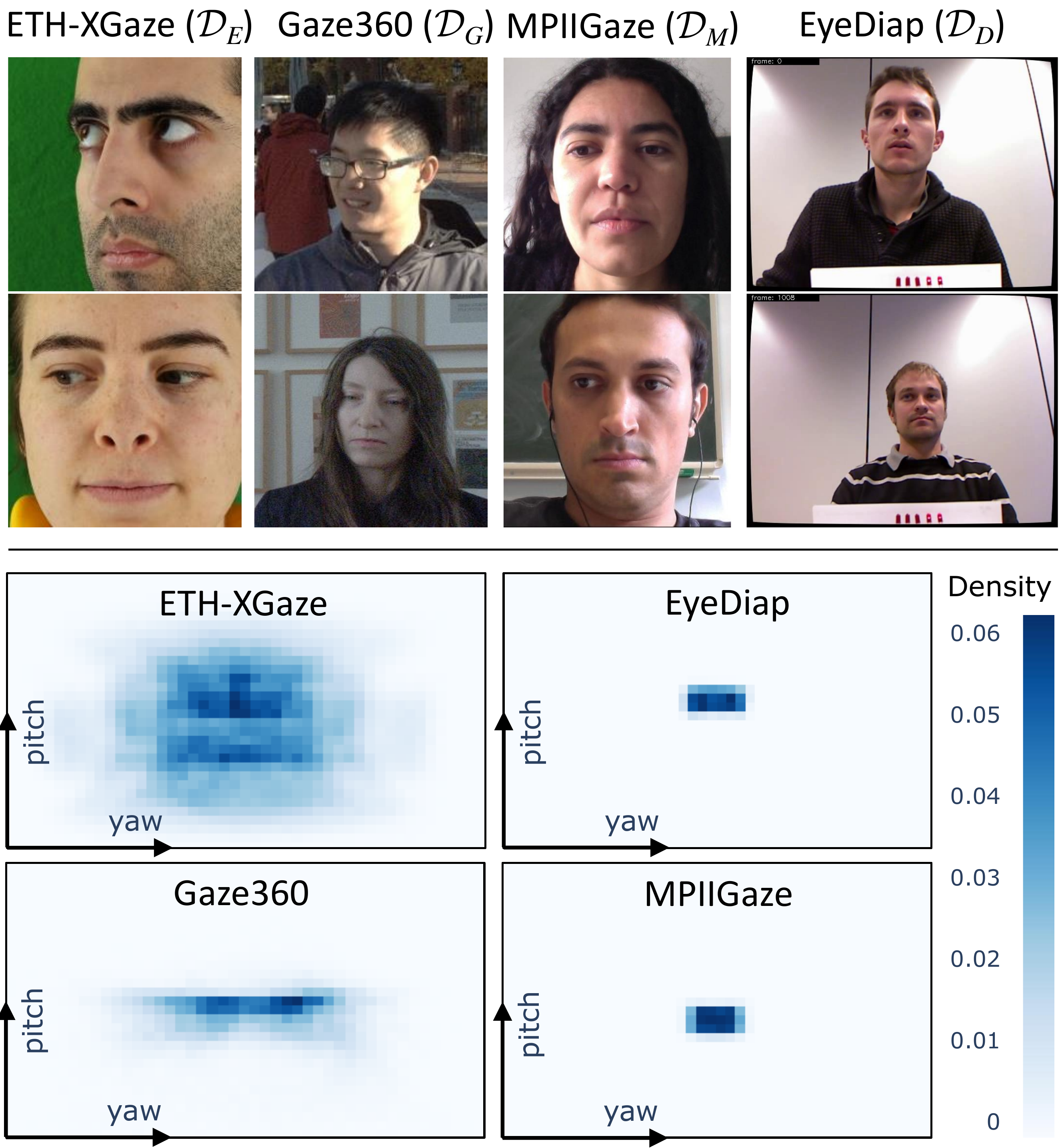}
	\end{center}
	\caption{Illustration of four gaze datasets. Top: Image samples. Large variation on illumination and background could be observed across datasets. Bottom: Gaze direction distributions. The x,y axes show the yaw and pitch angle.}
	\label{fig:samples}
\end{figure}

Image samples and gaze direction distributions of these datasets are shown in Fig. \ref{fig:samples}. 
\rev{By assuming the source domain has wider distributions than target domain,} 
we find that the gaze distributions of ETH-XGaze and Gaze360 are significantly larger than EyeDiap and MPIIGaze dataset, therefore we utilize ETH-XGaze and Gaze360 as source domains to pre-train models, and use MPIIGaze and EyeDiap as target domains for adaptation. It is noted that 1) we use gaze direction labels in ETH-XGaze and Gaze360 dataset for training the gaze estimation network and 2) labels in other datasets are only used for performance evaluation.

\section{Proposed Approach}

\begin{figure*}[htbp]
	\begin{center}
		\includegraphics[width=0.95\linewidth]{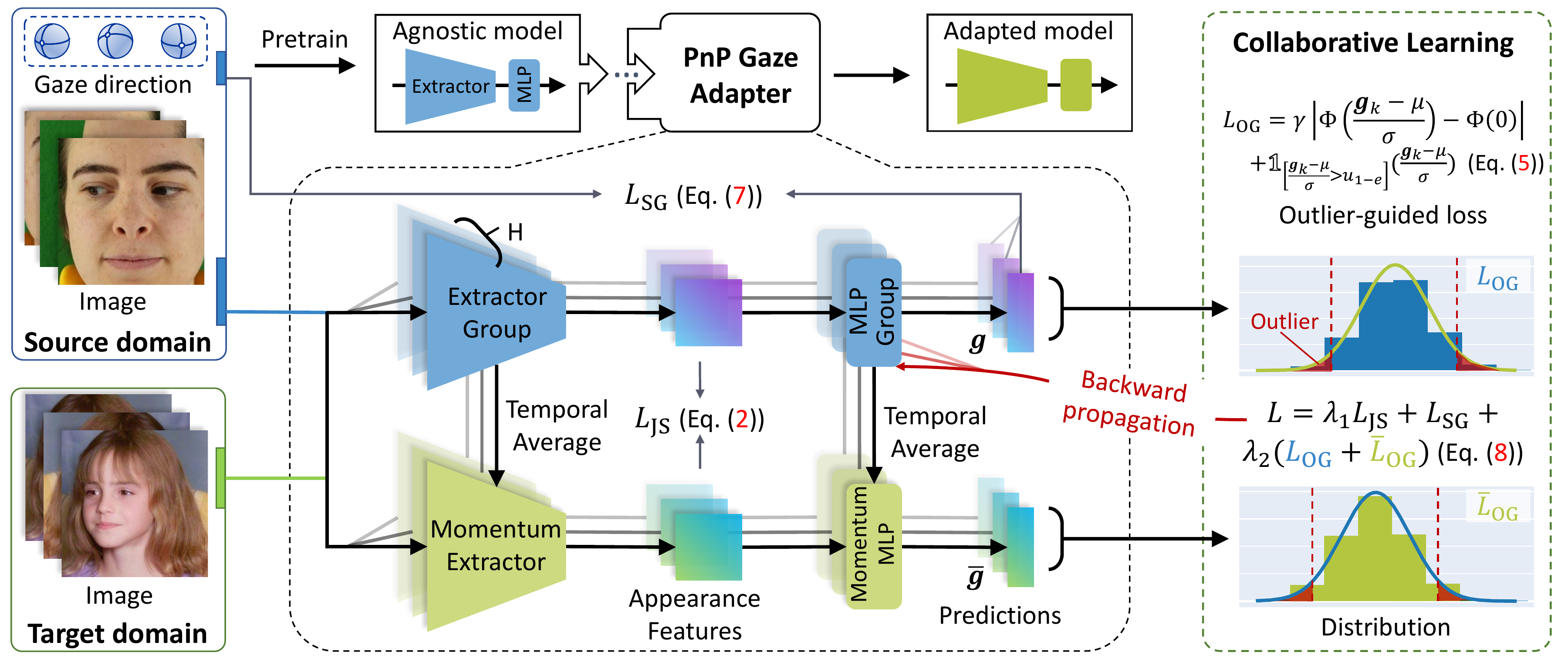}
	\end{center}
	\caption{Overview of the proposed \proposed~framework. 
	At first, a group of pre-trained models (with members' number $= H$) are plugged into our \proposed. 
	The architecture of \proposed~ is illustrated in the dark dashed box.
	Then a few samples from the source domain and the target domain are used for adaptation. 
	In the next, our \proposed~ collaboratively learns from both domains (Sec.~\ref{sec:collaborative}).
	During adaptation, an outlier-guided loss is proposed to better characterize the property of outliers (Sec~\ref{sec:og}).
	After adaptation, the first member of the model group is used as an adapted model, which achieves a better performance in the target domain.
	}
	\label{fig:overview}
\end{figure*}

We propose a novel Plug-and-Play Gaze Adaptation (\proposed) framework with outlier-guidance and collaborative learning. The architecture overview is illustrated in Fig.~\ref{fig:overview}. Our key idea is to guide the learning by the prediction outliers. We introduce a novel outlier-guided loss to better utilize the outliers for generalizing gaze estimation models. The proposed framework contains two branches, group $\mathbb{G}$ and group $\overline{\mathbb{G}}$, which collaboratively learn from each other. 
The \proposed~ framework is model agnostic and existing gaze estimation networks can be directly plugged into it without any modifications. 

\subsection{Model agnostic collaborative learning} \label{sec:collaborative}
The \proposed~framework employs two groups of networks that learn collaboratively. The group $\mathbb{G}$ generates gaze direction predictions with different pre-trained weights from the source domain. A network $G$ from $\mathbb{G}$ contains a feature extractor, which follows a multi-layer perceptron (MLP). While another group $\overline{\mathbb{G}}$ is the momentum version of group $\mathbb{G}$.
A network $\overline{G}$ from the momentum group $\overline{\mathbb{G}}$ takes the temporal average of the feature extractor and MLP from the corresponding network in group $\mathbb{G}$.
The \proposed ~framework is model agnostic since we use a generic network architecture, which consists of a feature extraction stage and a feature mapping stage. Such an architecture has been widely used in gaze estimation studies~\cite{G:chen2018dilated,G:cheng2020canet,G:zhang2017fullface}. 
Thus, different gaze estimation models can be easily plugged into the proposed \proposed~ framework.

We denote the feature extraction function of the \textit{k}-th network in group $\mathbb{G}$ as $F(\centerdot|\theta_k)$, the multi-layer perceptron of the \textit{k}-th network in group $\mathbb{G}$ as $\Pi(\centerdot|\theta_k)$. 
Then, the \textit{k}-th network $G(\centerdot|\theta_k)$ in group $\mathbb{G}$ can be expressed as $F(\centerdot|\theta_k)\circ \Pi(\centerdot|\theta_k)$. One straightforward way to generalize these networks to a target domain is to directly compute a loss from the outputs of themselves. 
However, the estimation errors might amplify during generalizing this way. To overcome error amplification, we propose to use the temporal average of these models to generate reliable appearance features and predictions. Specifically, the parameters of the \textit{k}-th model from the temporal average group at current iteration $T$ are denoted as $E^{(T)}[\theta_k]$, which can be updated as:

\begin{equation} \label{eq:tem}
	E^{(T)}[\theta_k]=\alpha E^{(T-1)}[\theta_k] + (1-\alpha)\theta_k,
\end{equation}
where $E^{(T-1)}[\theta_k]$ indicates the temporal average parameters of the \textit{k}-th network in the previous iteration $T-1$, and $\alpha$ represents the ensembling momentum, which is usually set as 0.99~\cite{D:ge2020mutual,D:he2019moco}. We utilize the robust appearance feature $\overline{z_k}$, which is generated by the \textit{k}-th momentum network as $\overline{F}(x|E[\theta_k])$ to constrain the \textit{k}-th appearance feature $z_k$, where $E[\theta_k]$ is the temporal average parameters of $\theta_k$. 
Specifically, we follow the work in~\cite{D:zhang2018deep} and minimize the symmetric Jensen-Shannon (JS) divergence between $\overline{z_k}$ and $z_k$ as follow:

\begin{equation} \label{eq:js}
	L_{\mathrm{JS}} = \frac{1}{2}(L_{\mathrm{KL}}(z_k\parallel \overline{z_k}) + L_{\mathrm{KL}}(\overline{z_k}\parallel z_k)),
\end{equation}
where $L_{\mathrm{KL}}(p\parallel q)$ is the Kullback Leibler (KL) divergence from $q$ to $p$, which is computed as:
\begin{equation} \label{eq:kl}
	L_{\mathrm{KL}}(p\parallel q) = \sum_t \log\frac{p(t)}{q(t)}.
\end{equation}

\begin{figure}[t]
	\begin{center}
		\includegraphics[width=\linewidth]{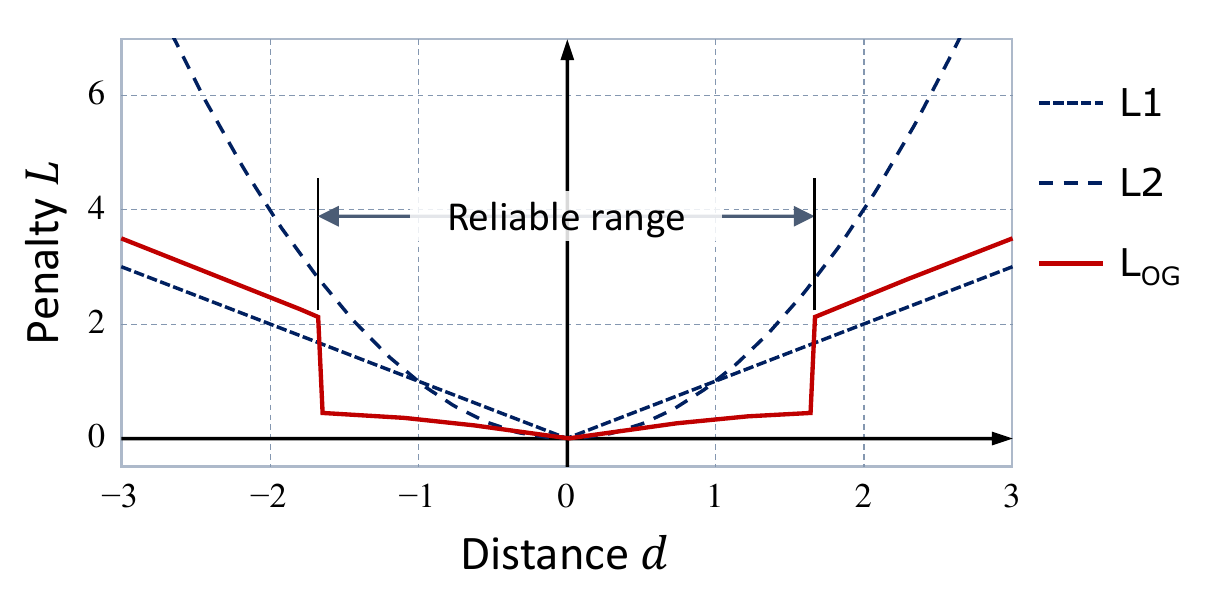}
	\end{center}
	\vspace{-2mm}
	\caption{Comparison of the proposed outlier-guided loss $L_{\mathrm{OG}}$, $L1$ loss, and $L2$ loss.}
	\label{fig:loss}
	\vspace{-5mm}
\end{figure}

\subsection{Outlier-guided loss function} \label{sec:og}
We specially design a loss function to better characterize the property of outliers. Suppose we have $H$ members in group $\mathbb{G}$ and group $\overline{\mathbb{G}}$ respectively, as illustrated in Fig.~\ref{fig:overview}. These networks output predictions $\mathbf{g}_k = G(\mathbf{x}|\theta_k)$, $\mathbf{\overline{g}}_k = \overline{G}(\mathbf{x}|E[\theta_k])$, where $k=1,...,H$. 
As mentioned in Sec.~\ref{sec:collaborative}, we make these two groups of networks guide each other to avoid error amplification.
Specifically, we calculate a normal distribution by estimating the mean $\mu$ and variance $\sigma^2$ of the momentum group's predictions using the following equations:
\begin{equation} \label{eq:mu-sigma}
	\mu = \frac{1}{H}\sum_{k=1}^{H} \mathbf{\overline{g}}_k, ~~ \sigma^2=\frac{1}{H - 1}\sum_{k=1}^H(\mathbf{\overline{g}}_k-\mu)^2.
\end{equation}

Here we define the output $\mathbf{g}_k$ as a reliable output when $|\frac{\mathbf{g}_k-\mu}{\sigma}| \leq u_{1-\epsilon}$, where $u$ is the quantile of the standard normal distribution, and $\epsilon$ indicates the significance level of judging an output as an outlier.
Then we estimate outliers from group $\mathbb{G}$'s predictions and punish them with a larger gradient. Toward this end, the outlier-guided loss is formulated as follow:
\begin{equation} \label{eq:loss-og}
	\begin{aligned}
		L_{\mathrm{OG}}(\mathbf{g}_k, \mu, \sigma)&=\gamma |\Phi(\frac{\mathbf{g}_k-\mu}{\sigma})-\Phi(0)|\\
		&+\mathbbm{1}_{[|\frac{\mathbf{g}_k-\mu}{\sigma}|>u_{1-\epsilon}]}|\frac{\mathbf{g}_k-\mu}{\sigma}|,
	\end{aligned}
\end{equation}
where $\Phi$ represents the cumulative distribution function of the standard normal distribution, \ie,
\begin{equation} \label{eq:phi}
	\begin{split}
		\Phi(x) = \int_{-\infty}^{x}\frac{1}{\sqrt{2\pi}}e^{-\frac{t^2}{2}}\text{dt}.
	\end{split}
\end{equation}
For better understanding, we visualize different loss functions of $L1$, $L2$ and $L_{\mathrm{OG}}$ in Fig.\ref{fig:loss}, in which distance $d=\frac{\mathbf{g}_k-\mu}{\sigma}$. In our experiments, we empirically choose $\gamma=0.01$, $\epsilon=0.05$. Please refer to Sec.~\ref{sec:hyperparam} for the parameter choosing. Similarly, we compute $\overline{L}_{\mathrm{OG}}$ for group $\overline{\mathbb{G}}$'s predictions.

\textbf{Source domain constraint.}
Besides the information obtained from the target domain, we find it also important to fully exploit the gaze information from the source domain. The gaze information of the source domain is learned by minimizing the distance between the model predictions $\mathbf{g}^s$ and the ground-truth labels $\mathbf{y}^s$, \ie,
\begin{equation} \label{eq:src}
	L_{\mathrm{SG}}(\mathbf{g^s}, \mathbf{y}^s)= |\mathbf{g^s} - \mathbf{y}^s|,
\end{equation}

\textbf{Total loss.} 
The total loss of the proposed system is a combination of several loss functions, which can be written as follow:

\begin{equation} \label{eq:total}
	L = \lambda_1 L_{\mathrm{JS}} + L_{\mathrm{SG}} + \lambda_2 ( L_{\mathrm{OG}} + \overline{L}_{\mathrm{OG}}),
\end{equation}
where $\lambda_1$ and $\lambda_2$ are tunable parameters. We emprically set $\lambda_1=0.01$ and $\lambda_2=0.1$ in our experiment.

\subsection{Adaptation procedure}

We summarize the adaptation procedure in Algorithm~\ref{alg:uma}. We first obtain a group of pre-trained networks $\mathbb{G}$, which contains the top \textit{H} networks trained on the source domain. The top \textit{H} networks are selected based on the prediction accuracy. We use a few images with ground-truth labels ($\mathcal{D}_s'$) from the source domain $\mathcal{D}_s$ and a few images ($\mathcal{D}_t'$) from the target domain $\mathcal{D}_t$ for unsupervised adaptation.
During the adaptation process, the networks in group $\mathbb{G}$ are trained by minimizing the loss function in Eq.~\eqref{eq:total}, and the momentum networks in group $\overline{\mathbb{G}}$ is updated with Eq.~\eqref{eq:tem}.After adaptation, the first member of the model group is used as an adapted model.

\begin{algorithm}[t]
	\caption{Outlier-guided domain adaptation algorithm for gaze estimation.}
	\label{alg:uma}
	
	\begin{algorithmic}[1]
		\Require {Small $\mathcal{D}_t'$, small $\mathcal{D}_s'$ and $\mathbb{G}$ pre-trained on $\mathcal{D}_s$}
		\Ensure{$G(\centerdot|\theta)$}
		
		\State Initialize: $\overline{\mathbb{G}}^{(1)} \gets \mathbb{G}^{(1)}$  \Comment{$\overline{\mathbb{G}}$ is the momentum $\mathbb{G}$.}

		\For{$ T{\gets}1\; to\; N $}
		
		\State ($\mathbf{x}_s$, $\mathbf{y}_s$),\; $\mathbf{x}_t$ $\gets$ $\mathcal{D}_s'$,\; $\mathcal{D}_t'$
		
		\State $L_{JS}$ $\gets$ $\mathbb{G}(\mathbf{x_t}|\theta^{(T)})$,\; $\overline{\mathbb{G}}(\mathbf{x_t}|E^{(T)}[\theta])$ with \eqnref{eq:js}.
		
		\State $L_{\mathrm{OG}}$, $\overline{L}_{\mathrm{OG}}$ $\gets$ $\mathbb{G}(\mathbf{x_t}|\theta^{(T)})$, $\overline{\mathbb{G}}(\mathbf{x_t}|E^{(T)}[\theta])$ with \eqnref{eq:loss-og}
		
		\State $L_{\mathrm{SG}}$ $\gets$ $\mathbf{y}_s$,\; $\mathbb{G}(\mathbf{x}_s|\theta^{(T)})$ with \eqnref{eq:src}.
		
		\State Train $\mathbb{G}(\centerdot|\theta^{(T+1)})$ with \eqnref{eq:total}.
		
		\State Update $ E^{(T+1)}[\theta]$ in $\overline{\mathbb{G}}$ with \eqnref{eq:tem}
		
		\EndFor
		\State $G(\centerdot|\theta) \gets \mathbb{G}(\centerdot|\theta)_1$
	\end{algorithmic}
\end{algorithm}

\textbf{Training details.} 
We employ PyTorch for implementation. All experiments run on a single NVIDIA 2080TI GPU. We use the Adam optimizer with a learning rate of $10^{-4}$ to train the gaze estimation network in the source domain. For model adaptation, we use the Adam optimizer with a learning rate of $10^{-4}$. 

\section{Experimental Results}

\subsection{Data preparation}

Due to the differences in different dataset, we first do data preparation. Note that our domain adaptation task is to adapt from the source domain (ETH-XGaze or Gaze360 dataset) to the target domain (MPIIGaze or EyeDiap dataset). 
For the ETH-XGaze dataset~\cite{G:zhang2020eth} ($\mathcal{D}_E$), it provides 80 subjects (\ie, 756,540 images) which are used as the training set. For the Gaze360 dataset~\cite{G:kellnhofer2019gaze360} ($\mathcal{D}_G$), we remove images without subjects' faces and use the remaining 100,933 images as the training set. Since the MPIIGaze\cite{G:zhang2017mpiigaze} ($\mathcal{D}_M$) provides a standard evaluation protocol, which selects 3000 images from each subject to form an evaluation set, therefore we adopt the evaluation set directly. For the EyeDiap dataset~\cite{G:funes2014eyediap} ($\mathcal{D}_D$), we employ 16,674 images from 14 subjects under screen target sessions as the evaluation set.

We follow the approach in~\cite{G:zhang2017mpiigaze} to normalize the data. They predict the gaze direction only based on face images. However, the change of head poses will significantly affect the appearance of face image. Thus, we eliminated the influence of different head poses through rotating the virtual camera and wrapping the images.

\begin{table*}[htbp]
\caption{Domain adaptation results of plugging the existing gaze estimation networks into the proposed \proposed~framework. Angular gaze error ($^{\circ}$) is used as evaluation metric.} \label{tab:backbones}
\vspace{-2mm}
	\begin{center}
		\begin{tabular}{clcccc}
			\toprule[1.2pt]
			Input & Method & $\mathcal{D}_E\rightarrow\mathcal{D}_M$ & $\mathcal{D}_E\rightarrow\mathcal{D}_D$ & $\mathcal{D}_G\rightarrow\mathcal{D}_M$ & $\mathcal{D}_G\rightarrow\mathcal{D}_D$ \\
			
			\hline
			\specialrule{0em}{1pt}{1pt}
			
			\multirow{6}{*}{\rotatebox{90}{Face}}
			
			& Baseline~\cite{O:he2016resnet}        & 8.767 & 8.578 & 7.662 & 8.977 \\
			& Baseline~\cite{O:he2016resnet} + \proposed & \textbf{5.529} \upscore{36.9} & \textbf{5.867} \upscore{31.6} & \textbf{6.176} \upscore{19.4} & \textbf{7.922} \upscore{11.8} \\\cline{2-6} \specialrule{0em}{1pt}{1pt}
			
			& ResNet50~\cite{O:he2016resnet,G:zhang2020eth}        & 8.017 & 8.310 & 8.328 & 7.549 \\
			& ResNet50~\cite{O:he2016resnet,G:zhang2020eth} + \proposed & \textbf{6.000} \upscore{25.2} & \textbf{6.172} \upscore{25.7} & \textbf{5.739} \upscore{31.1} & \textbf{7.042} \upscore{6.7} \\\cline{2-6} \specialrule{0em}{1pt}{1pt}
			
			& SWCNN~\cite{G:zhang2017fullface}        & 10.939 & 24.941 & 10.021 & 13.473 \\
			& SWCNN~\cite{G:zhang2017fullface} + \proposed & \textbf{8.139} \upscore{25.6} & \textbf{15.794} \upscore{36.7} & \textbf{8.740} \upscore{12.8} & \textbf{11.376} \upscore{15.6} \\
			
			\hline
			\specialrule{0em}{1pt}{1pt}
			
			\multirow{4}{*}{\rotatebox{90}{Face + Eye}}
			
			& CA-Net~\cite{G:cheng2020canet}        & - & - & 21.276 & 30.890 \\
			& CA-Net~\cite{G:cheng2020canet} + \proposed & - & - & \textbf{17.597} \upscore{17.3} & \textbf{16.999} \upscore{44.9} \\\cline{2-6} \specialrule{0em}{1pt}{1pt}
			
			& Dilated-Net~\cite{G:chen2018dilated}        & - & - & 16.683 & 18.996 \\
			& Dilated-Net~\cite{G:chen2018dilated} + \proposed & - & - & \textbf{15.461} \upscore{7.3} & \textbf{16.835} \upscore{11.4} \\
			\bottomrule[1.2pt]
		\end{tabular}
	\end{center}
	\vspace{-6mm}
\end{table*}

\subsection{Plugging the existing gaze estimation networks}

We first plug the existing gaze estimation networks into the \proposed~ framework to improve their domain adaptation performances. We select several popular gaze estimation networks: ResNet18~\cite{O:he2016resnet}, ResNet50~\cite{O:he2016resnet,G:zhang2020eth}, SWCNN~\cite{G:zhang2017fullface}, CA-Net\cite{G:cheng2020canet} and Dilated-Net~\cite{G:chen2018dilated}. The baseline method is ResNet18~\cite{O:he2016resnet}.
Two different types of networks are utilized to evaluate our method: 1) ResNet18, ResNet50 and SWCNN only require facial images as input, and 2) CA-Net and Dilated-Net require both face and eye images as input. It is noted that since ETH-XGaze dataset provides only face images, it cannot be used as the source domain for CA-Net or Dialated-Net.

Table~\ref{tab:backbones} shows the result of plugging the existing networks into our proposed method. Due to the variety in the scale of different backbones, except the baseline (ResNet18), 3 pre-trained models are used to form $\mathbb{G}$ and $\overline{\mathbb{G}}$. 10 images from the target domain are used to conduct the experiments. With our \proposed~ framework, numerical results show that all gaze estimation networks achieve significant improvement on these four domain adaptation tasks.
Quantitative results demonstrate that the \proposed~ framework achieves considerable performance improvements of 36.9\%, 31.6\%, 19.4\% and 11.8\% over the baseline system on ETH-to-MPII, ETH-to-EyeDiap, Gaze360-to-MPII and Gaze360-to-EyeDiap tasks.

\subsection{Comparison with state-of-the-art domain adaptation approaches}

We compare the cross-dataset performance of our \proposed~framework with the state-of-the-art methods on unsupervised gaze domain adaptation.

\begin{itemize}
	\item Source only : Inspired by \cite{G:zhang2020eth}, we use ResNet18\cite{O:he2016resnet} as the baseline of our gaze estimation network, which has no domain adaptation.
	\item DAGEN \cite{G:guo2020dagaze}: A state-of-the-art unsupervised domain adaptation method proposed for person specific gaze estimation within dataset. 500 target samples are used for adaptation to reach a better performance.
	\item UMA \cite{D:cai2020uma}: A state-of-the-art unsupervised domain adaptation method proposed for hand segmentation, which uses self-training to decrease the domain gap. We modify it slightly to fit our task. 100 target samples are used for adaptation to reach a better performance.
	\item ADDA \cite{D:tzeng2017adversarial}: An unsupervised domain adaptation method, which uses adversarial learning to decrease the domain gap. 500 target samples are used for adaptation to reach a better performance.
	\item GVBGD \cite{D:cui2020gradually}: A state-of-the-art unsupervised domain adaptation method. It equips adversarial domain adaptation with gradually vanishing bridge mechanism on both generator and discriminator. 1000 target samples are used for adaptation to reach a better performance.
\end{itemize}

ADDA~\cite{D:tzeng2017adversarial} and GVBGD~\cite{D:cui2020gradually} are originally proposed for classification and UMA~\cite{D:cai2020uma} is originally proposed for hand segmentation. We compare these methods here to show how the state-of-the-art domain adaptation methods could help to improve the generalization performance of gaze estimation. To make a fair comparison, their original networks are replaced with ResNet18, which is the same as our baseline.

Quantitative results of different methods are shown in Table \ref{tab:compare}. Our method significantly outperforms the state-of-the-art domain adaptation method. 
The superior performance of our method over DAGEN~\cite{G:guo2020dagaze}, UMA~\cite{D:cai2020uma}, ADDA~\cite{D:tzeng2017adversarial} and GVBGD~\cite{D:cui2020gradually} verifies the effectiveness of the proposed \proposed~ for generalizing gaze estimation.

\newcommand{\tabincell}[2]{\begin{tabular}{@{}#1@{}}#2\end{tabular}}  
\begin{table}[htbp]
\caption{Comparison with the state-of-the-art domain adaptation approaches. Angular gaze error ($^{\circ}$) is used as evaluation metric. * indicates that more than 100 images from the target domain are needed for adaptation. ** denotes that gaze direction labels in the target domain are needed.}
\label{tab:compare}
\vspace{-3mm}
	\begin{center}
	\small
	\setlength{\tabcolsep}{0.5mm}{
		\begin{tabular}{lcccc}
		\toprule[1.2pt]
			Method & $\mathcal{D}_E \rightarrow$ $\mathcal{D}_M$ & $\mathcal{D}_E \rightarrow$ $\mathcal{D}_D$ & $\mathcal{D}_G \rightarrow$  $\mathcal{D}_M$ & $\mathcal{D}_G \rightarrow$ $\mathcal{D}_D$ \\
			\hline \specialrule{0em}{1pt}{1pt}
			Source only  & 8.77 & 8.58 & 7.66 & 8.98 \\
			Fine-tune ** & 7.14 & 8.25 & 8.55 & 8.95\\
			\hline \specialrule{0em}{1pt}{1pt}
			Gaze360 \cite{G:kellnhofer2019gaze360} * & 5.97 & 7.84 & 7.38 & 9.61 \\		
			GazeAdv \cite{G:wang2019adgaze} *  		 & 6.75 & 8.10 & 8.19 & 12.27 \\	
			DAGEN \cite{G:guo2020dagaze} * 			 & 6.16 & 9.73 & 6.61 & 12.90 \\
			ADDA \cite{D:tzeng2017adversarial} *  	 & 6.33 & 7.90 & 8.76 & 14.80 \\
			GVBGD \cite{D:cui2020gradually} * 		 & 6.68 & 7.27 & 7.64 & 12.44 \\
			UMA \cite{D:cai2020uma} 				 & 7.52 & 12.37 & 8.51 & 19.32 \\
			Ours & \textbf{5.53} & \textbf{5.87} & \textbf{6.18} & \textbf{7.92}\\
		\bottomrule[1.2pt]
		\end{tabular}
	}
	\end{center}
	\vspace{-7mm}
\end{table}

\subsection{System characteristics analysis}

\subsubsection{Ablation study}

\begin{table}
\caption{Ablation study results: cross-dataset gaze estimation performance. Angular gaze error ($^{\circ}$) is used as evaluation metric.}
\label{tab:abla}
\vspace{-3mm}
    \setlength{\tabcolsep}{2.5mm}{
	\begin{center}
		\begin{tabular}{lcc}
		\toprule[1.2pt]
			Method & $\mathcal{D}_E \rightarrow$  $\mathcal{D}_M$ & $\mathcal{D}_G \rightarrow$ $\mathcal{D}_M$ \\
			\hline  \specialrule{0em}{1pt}{1pt}
			Baseline (Source only) & 8.77 & 7.66 \\
			\hline  \specialrule{0em}{1pt}{1pt}
            CNN+oma                & $5.79_{\pm0.19}$ & $7.01_{\pm0.42}$ \\
            CNN+2oma               & $5.65_{\pm0.16}$  & $6.86_{\pm0.50}$\\
            CNN+oma+js             & $5.63_{\pm0.12}$ & $6.62_{\pm0.29}$\\
            CNN+2oma+js            & $5.62_{\pm0.22}$ & $6.51_{\pm0.42}$\\
            CNN+2oma+js+sg (final) & $\textbf{5.53}_{\pm0.24}$ & $\textbf{6.18}_{\pm0.41}$ \\
        
        \bottomrule[1.2pt]
		\end{tabular}
	\end{center}}
	\vspace{-10mm}
\end{table}

We conduct an ablation study to test the effectiveness of different components in the proposed method. The components are shown below.

\begin{itemize}
	\item Baseline: a standard CNN-based gaze estimation model using architecture of ResNet18\cite{O:he2016resnet}. The model is pretrained on the source domain.
	\item oma: outlier-guided model adaptation (\eqnref{eq:loss-og}) in which the normal distribution is estimated based on the off-line predictions (\ie, the outputs of $\overline{\mathbb{G}}$).
	\item 2oma: outlier-guided model adaptation (\eqnref{eq:loss-og}) in which two normal distributions are estimated based on the on-line and the off-line predictions, respectively.
	\item js: symmetric Jensen-Shannon divergence (\eqnref{eq:js}), which is used to constrain the distance of feature distributions between $\mathbb{G}$ and $\overline{\mathbb{G}}$.
	\item sg: gaze constraint from the source domain (\eqnref{eq:src}).
\end{itemize}

For all the experiments, we set the number of group members as 10. These pre-trained models are fixed as initial parameters for all the ablation studies.
The choice verification results are shown in Sec.~\ref{sec:models}. Table \ref{tab:abla} shows the gaze estimation errors under different combinations. We observe that the outlier-guided model adaptation (CNN+oma) can significantly improve the gaze estimation performance over the source only approach. Adding symmetric Jensen-Shannon divergence (js) and on-line outlier-guided adaptation (2oma) further improves the gaze estimation performance, and our final version achieves the best results on all tasks, which confirms our hypothesis that inputting the source domain information helps the models better learn domain invariant information.

\subsubsection{Loss functions}

To verify the effectiveness of our proposed outlier-guided loss, we compare it with L1 and L2 loss that are commonly used on regression tasks. The final version in Table~\ref{tab:abla} is used as the basic structure, and the results are shown in Table \ref{tab:losses}.

\begin{table}[htbp]
\caption{Performance comparison of different loss functions. Angular gaze error ($^{\circ}$) is used as evaluation metric.}
\label{tab:losses}
\setlength{\tabcolsep}{1mm}{
	\begin{center}
		\begin{tabular}{lcccc}
			\toprule[1.2pt]
			Loss & $\mathcal{D}_E\rightarrow\mathcal{D}_M$ & $\mathcal{D}_E\rightarrow\mathcal{D}_D$ & $\mathcal{D}_G\rightarrow\mathcal{D}_M$ & $\mathcal{D}_G\rightarrow\mathcal{D}_D$ \\
			\hline
			$L1$ & $7.00_{\pm0.76}$ & $6.22_{\pm0.19}$ & $7.03_{\pm0.25}$ & $8.24_{\pm0.11}$ \\
			$L2$ & $6.00_{\pm0.22}$ & $5.96_{\pm0.14}$ & $6.46_{\pm0.21}$ & $8.23_{\pm0.16}$ \\
			$L_{\mathrm{OG}}$ & $\textbf{5.53}_{\pm0.24}$ & $\textbf{5.87}_{\pm0.16}$ & $\textbf{6.18}_{\pm0.41}$ & $\textbf{7.92}_{\pm0.15}$ \\
			\bottomrule[1.2pt]
		\end{tabular}
	\end{center}}

\end{table}

The results show that the proposed outlier-guided loss consistently outperforms the L1 and L2 loss on all four tasks. Since the predictions are mostly lie in the range of $[-0.5,0.5]$, we find that the system performance degrades when the slope of the loss function increases.

\subsubsection{Hyperparameters} \label{sec:hyperparam}

We evaluate how the \proposed~ performance varies with the change of significance level $\epsilon$ (Eq. \ref{eq:loss-og}). 
$\epsilon$ controls the range of reliable range of the predictions, which is also illustrated in Fig.~\ref{fig:loss}.
We test three significance levels which are commonly used in statistical analysis, 0.1, 0.05, and 0.01. 
The results are shown in row 1-3 of Table \ref{tab:sigs}, where we find that the best system performance occurs at the significance level of $\epsilon=0.05$. 

\begin{table}[htbp]
\caption{Cross-dataset gaze estimation performance of different significance levels $\epsilon$ and weight parameters $\gamma$. Angular gaze error ($^{\circ}$) is used as evaluation metric.}
\label{tab:sigs}
\setlength{\tabcolsep}{0.5mm}{
	\begin{center}
		\begin{tabular}{lcccc}
			\toprule[1.2pt]
			  & $\mathcal{D}_E\rightarrow\mathcal{D}_M$ & $\mathcal{D}_E\rightarrow\mathcal{D}_D$ & $\mathcal{D}_G\rightarrow\mathcal{D}_M$ & $\mathcal{D}_G\rightarrow\mathcal{D}_D$ \\
			\hline
			$\epsilon=0.1$ & $6.11_{\pm0.51}$ & $5.91_{\pm0.07}$ & $6.79_{\pm0.59}$ & $7.93_{\pm0.30}$ \\
			$\epsilon=0.05$ & $\textbf{5.53}_{\pm0.24}$ & $\textbf{5.87}_{\pm0.16}$ & $\textbf{6.18}_{\pm0.41}$ & $\textbf{7.92}_{\pm0.15}$ \\
			$\epsilon=0.01$ & $5.94_{\pm0.54}$ & $6.00_{\pm0.17}$ & $6.97_{\pm0.52}$ & $7.93_{\pm0.32}$ \\
			\hline
			$\gamma=1$ & $5.62_{\pm0.46}$ & $5.87_{\pm0.29}$ & $7.23_{\pm0.91}$ & $7.99_{\pm0.42}$ \\
			$\gamma=0.1$ & $5.79_{\pm0.38}$ & $5.88_{\pm0.24}$ & $6.76_{\pm0.56}$ & $7.97_{\pm0.17}$ \\
			$\gamma=0.01$ & $\textbf{5.53}_{\pm0.24}$ & $\textbf{5.87}_{\pm0.16}$ & $\textbf{6.18}_{\pm0.41}$ & $\textbf{7.92}_{\pm0.15}$ \\
			$\gamma=0.001$ & $5.76_{\pm0.47}$ & $6.50_{\pm0.49}$ & $6.61_{\pm0.52}$ & $7.99_{\pm0.27}$ \\
			\bottomrule[1.2pt]
		\end{tabular}
	\end{center}}
\end{table}

Furthermore, we evaluate how the estimation performance varies with weight parameter $\gamma$ (Eq. \ref{eq:loss-og}). We conduct experiments with different $\gamma$ on all four tasks. The results are shown in Table~\ref{tab:sigs}. It indicates that the proposed \proposed~ reaches the best performance at the value $\gamma=0.01$. 
Therefore, we set $\epsilon=0.05$ and $\gamma=0.01$ for the remaining experiments.


\subsubsection{Number of group members} \label{sec:models}
\begin{figure}[t]
	\begin{center}
		\includegraphics[width=0.9\linewidth]{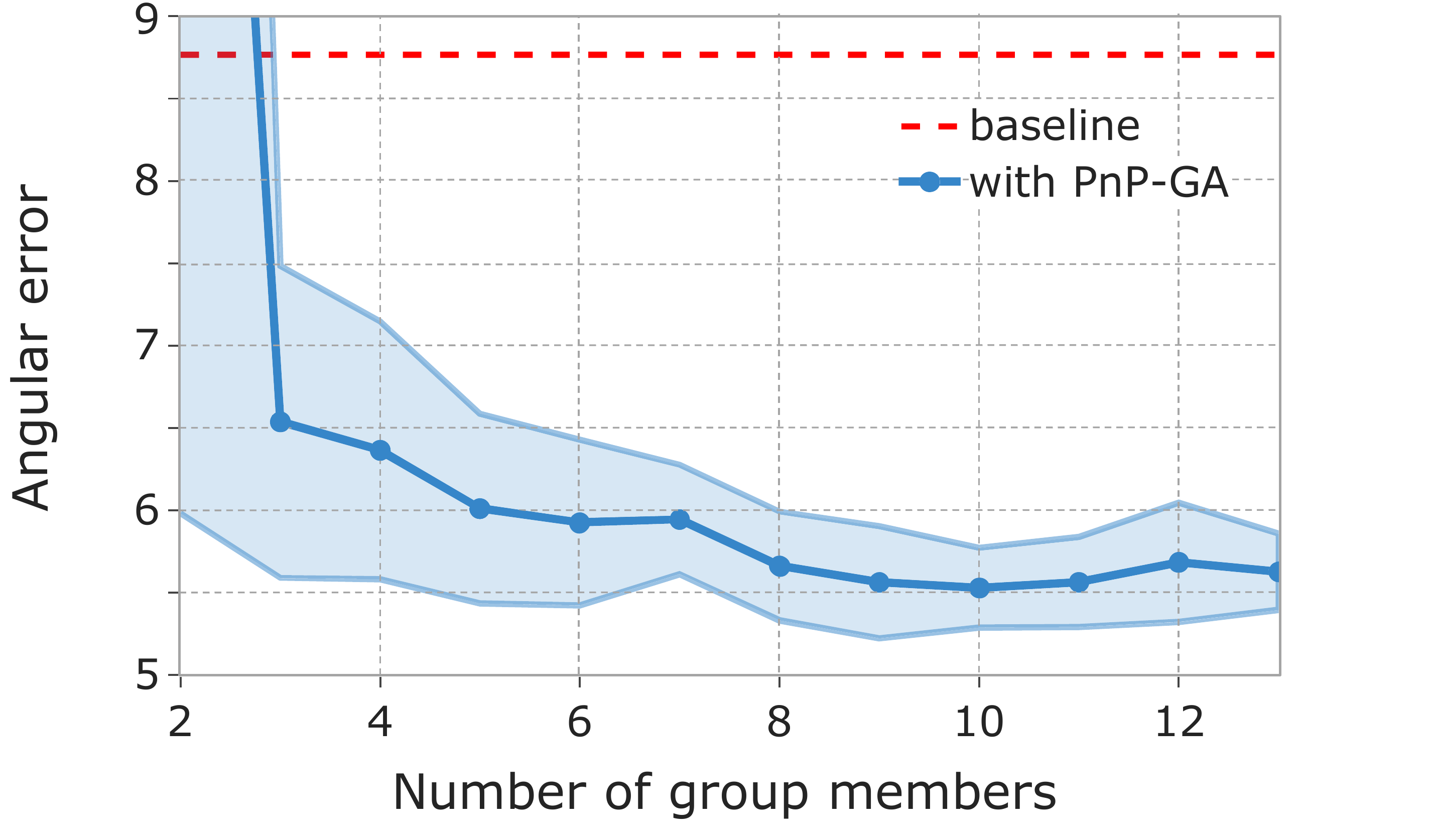}
	\end{center}
	\caption{Evaluation of model adaptation performance with gradually increasing the number of group members.}
	\label{fig:models}
	\vspace{-4mm}
\end{figure}

To find the optimal number of group members (\ie, $H$ in Fig. \ref{fig:overview}), we evaluate the system performance under different numbers of group members on $\mathcal{D}_E\rightarrow\mathcal{D}_M$ task.
In detail, we run our method under five different random seeds for different $H$. 
The average performance and variance are calculated for each run.
The results are illustrated in Fig. \ref{fig:models}. 
It indicates that the performance of \proposed~ constantly improves as the number of group members increases. The system performance converges when $H\geq10$. 
Consequently, we choose $H=10$ for our \proposed.

\begin{figure}
	\centering
	\includegraphics[width=\linewidth]{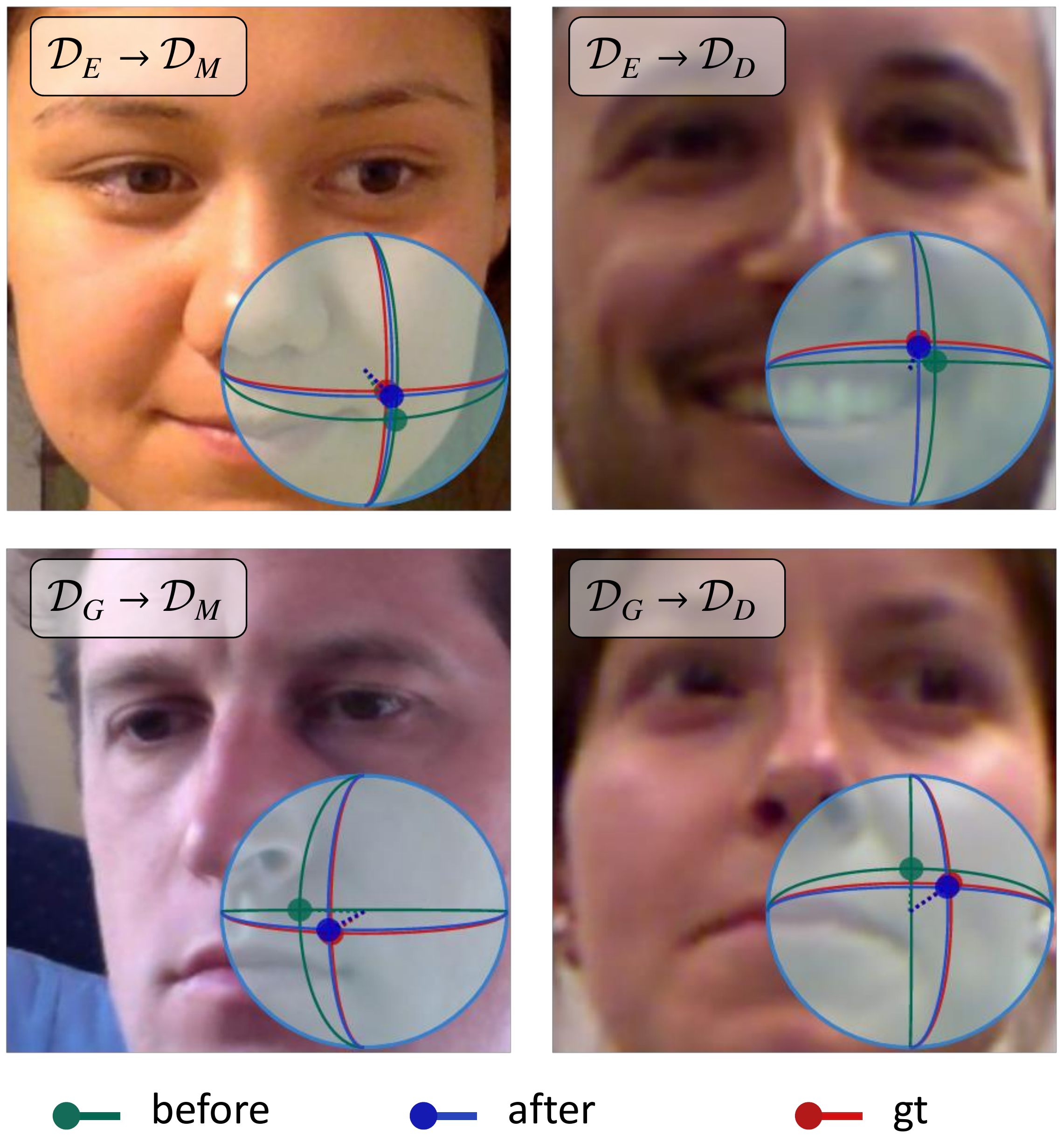}
	\caption{Visual results example of estimated 3D gaze. Red points represent the ground-truth gaze direction, green and blue points represent the predictions before and after adaptation, respectively. }
	\label{fig: visualgaze}
\end{figure}

\begin{figure}
	\centering
	\includegraphics[width=\linewidth]{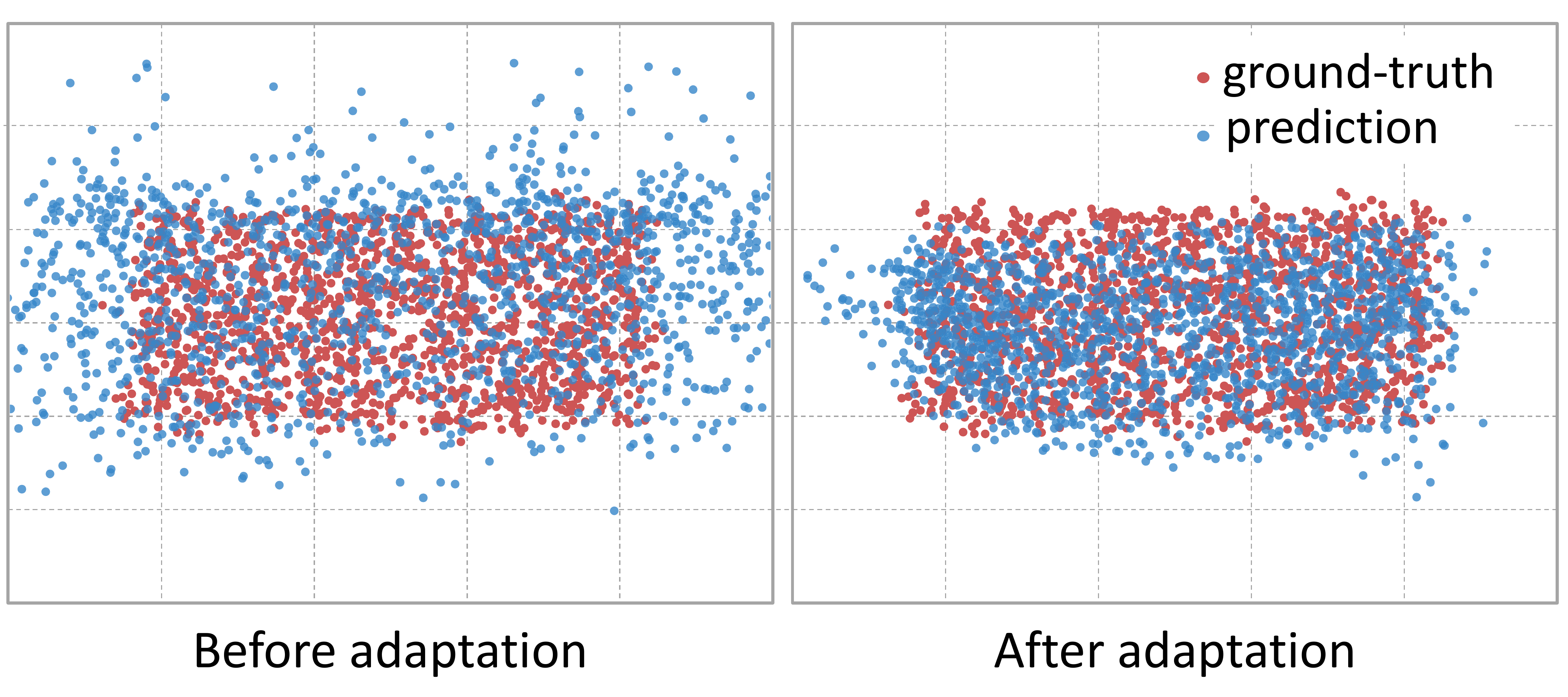}
	\caption{Scatter plot of predictions and ground-truth labels before and after model adaptation. Scatter points in blue represent predictions and red ones represent labels.}
	\label{fig: scatter}
\end{figure}

\subsection{Comprehension of \proposed} \label{sec:visual}
To understand how our method improves the performance of gaze estimation in target domains, 
we 1) visualize typical cases by drawing gaze predictions on the input and 
2) compare the prediction distributions before/after gaze domain adaptation. 
In detail, examples of visual gaze estimation results on four adaptation tasks are illustrated in Fig. \ref{fig: visualgaze}. 
Results show that our proposed \proposed~ can adapt well in different conditions. 
When the gaze direction deviates from the face direction, our method can also generate an accurate gaze direction (right bottom figure in Fig.~\ref{fig: visualgaze}).

Fig. \ref{fig: scatter} shows the distributions of predictions before and after adaptation. 
Our method significantly reduces the degree of outlier, the prediction distribution of our method are much closer to the distribution of ground-truth labels.
These results also reveal that how our \proposed~ improves the performance of gaze estimation.

\section{Conclusion}

In this paper we presented a novel outlier-guided plug-and-play gaze adaptation framework (\proposed) for generalizing the gaze estimation to new domains. 
The proposed \proposed~ framework is model agnostic, in which most existing gaze estimation networks can be plugged without any modifications. Collaborative learning process and a novel outlier-guided loss are proposed for \proposed. Our method demonstrates superior performance on four gaze domain adaptation tasks, as well as exceptionally improved performance of plugging existing gaze estimation networks into the \proposed~ framework.

{\small
\bibliographystyle{ieee_fullname}
\bibliography{egbib}
}

\end{document}